# 2kenize: Tying Subword Sequences for Chinese Script Conversion

**Pranav A**[徒] and **Isabelle Augenstein**[師]

[徒]Independent Researcher, Hong Kong
[師]Department of Computer Science, University of Copenhagen, Denmark
cs.pranav.a{at}gmail.com, augenstein{at}di.ku.dk

## Abstract

Simplified Chinese to Traditional Chinese character conversion is a common preprocessing step in Chinese NLP. Despite this, current approaches have poor performance because they do not take into account that a simplified Chinese character can correspond to multiple traditional characters. Here, we propose a model that can disambiguate between mappings and convert between the two scripts. The model is based on subword segmentation, two language models, as well as a method for mapping between subword sequences. We further construct benchmark datasets for topic classification and script conversion. Our proposed method outperforms previous Chinese Character conversion approaches by 6 points in accuracy. These results are further confirmed in a downstream application, where 2kenize is used to convert pretraining dataset for topic classification. An error analysis reveals that our method's particular strengths are in dealing with code mixing and named entities. The code and dataset is available at https://github.com/pranav-ust/2kenize

## 1 Introduction

Chinese character (or script) conversion is a common preprocessing step for Chinese NLP practitioners (Zhang, 2014; Shi et al., 2011). Traditional Chinese (TC) and Simplified Chinese (SC) are the two standardized character sets (or scripts) for written Chinese. TC is predominantly used in Taiwan, Hong Kong, and Macau, whereas SC is mainly adopted in mainland China and SC characters are simplified versions of TC characters in terms of strokes and parts. Therefore, Chinese NLP practitioners apply script converters[1] to translate the dataset into their desired language. This is especially useful for TC NLP practitioners because TC is less widely used and under-resourced as compared to SC.

| SC Sentence | 维护发展中国家共同利益 | | Comments |
|---|---|---|---|
| Segmentation | 维 护发 展 中 国家 共同 利益 | | 护发: *haircare* |
| Conversion | 維護髮展中國家共同利益 | | ✗ Conversion |
| Segmentation | 维 护 发展 中 国家 共同 利益 | | 发展: *develop* |
| Conversion | 維護發展中國家共同利益 | | ✓ Conversion |

Table 1: Example sentence with two different segmentations, and resulting different conversions. The sentence translates to *Safeguarding the common interests of developing countries*. This is a recurring example in this paper. Also refer §F.5.

Converting from TC to SC is generally straightforward because there are one-to-one correspondences between most of the characters, so conversion can be performed using mapping tables (Denisowski, 2019; Chu et al., 2012). However, conversion from SC to TC is an arduous task as some SC characters can be mapped to more than one TC character depending on the context of the sentence. A detailed analysis by Halpern and Kerman (1999) shows that SC to TC conversion is a challenging and crucial problem, as 12% of SC characters have one-to-many mappings to TC characters. Our experiments show that current script converters achieve sentence accuracy results of 55-85% (§3).

Another issue is that varying tokenization would lead to different results as Chinese is an unsegmented language, see Table 1 for an example. Off-the-shelf script converters would translate 维护发展中国家共同利益 into 維護髮展中國家共同利益[2], whereas the correct conversion is 維

---

[2]Throughout this paper, we color code ambiguous SC characters with brown, ambiguous TC characters with violet, vernacular Cantonese characters with teal. By scripts, we refer as to character sets, and we interchangeably use them in this paper.

護發展中國家共同利益. Here, the SC character 发 (*hair, issue*) has two TC mappings, 髮 (*hair, issue*) and 發 (*hair, issue*), depending on the context and tokenization; which shows that this task is non-trivial.

Despite this being an important task, there is a lack of benchmarks[3], which implies that this problem is understudied in NLP. In this study, we propose *2kenize*, a subword segmentation model which jointly considers Simplified Chinese and forecasting Traditional Chinese constructions. We achieve this by constructing a joint Simplified Chinese and Traditional Chinese language model based Viterbi tokenizer. Performing mapping disambiguation based on this tokenization method improves sentence accuracy by 6 points as compared to off-the-shelf converters and supervised models. Our qualitative error analysis reveals that our method's particular strengths are in dealing with code-mixing and named entities. Additionally, we address the issue of a lack of benchmark datasets by constructing datasets for script conversion and TC topic classification.

## 2   2kenize: Joint Segmentation and Conversion

We employ subword tokenization, as it addresses the issue of rare and unknown words (Mikolov et al., 2012) and has been shown advantageous for the language modelling of morphologically-rich languages (Czapla et al., 2018; Mielke and Eisner, 2019). This achieves improvements in accuracy for neural machine translation (NMT) tasks and has now become a prevailing practice (Denkowski and Neubig, 2017). The most widely-utilized method is Byte Pair Encoding (BPE, Sennrich et al. (2016)), a compression algorithm that combines frequent sequences of characters, which results in rare strings being segmented into subwords. Unigram (Kudo, 2018) and BPE-Drop (Provilkov et al., 2019) use subword ambiguity as noise, as well as stochastically-corrupted BPE segmentation to make it less deterministic. For NMT tasks generally, subword segmentation is seen as a monolingual task and applied independently on source and target corpora. We hypothesize that translation tasks, and specifically conversion tasks, as investigated here, would have a bet-ter performance if segmentation were performed jointly. Hence, in this section, we describe our proposed method *2kenize*, which jointly segments by taking the source and its approximate target sentences into account. This motivates the main idea of this paper: We propose *2kenize* which jointly considers the source sentence and its corresponding target conversions by doing lookaheads with mappings.

### 2.1   Outline of the proposed approach

Given the possible SC character sequence $\mathbf{s} = s_1 s_2 \ldots s_n$ and TC character sequence $\mathbf{t} = t_1 t_2 \ldots t_n$, we want to find the most likely $\mathbf{t}$, which is given by the Bayes decision rule as follows:

$$\mathbf{t} = \arg\max_{\mathbf{t}' \in T^*} p(\mathbf{s}, \mathbf{t}') \qquad (1)$$

where $T^*$ denotes the set of all strings over symbols ($t_i$) in $T$ (Kleene star). We divide this problem into two parts: finding the mapping sequence (2) and finding the TC sequence from mappings (7).

We define a mapping, which is given by $m_i = (\mathfrak{s}_i, \mathfrak{t}_i) = (s_{j:k}, \mathfrak{t}_{j:k})$. Here, $\mathfrak{t}_{j:k} = \{t_{j:k}^1 \ldots t_{j:k}^n\}$ is a set of TC characters that correspond to the SC character in the mapping. Thus, a mapping sequence can be defined as a concatenation of mappings, which is $\mathbf{m} = m_1 m_2 \ldots m_l$. Let $\mathcal{M}$ be the superset of all possible mapping sequences and $\mathcal{M}(\mathbf{s})$ be the all mapping sequences resulting from $\mathbf{s}$. Then, the best possible mapping sequence is given by

$$\mathbf{m} = \arg\max_{\mathbf{m}' \in \mathcal{M}(\mathbf{s})} p(\mathbf{m}') \qquad (2)$$

Morever, $p(\mathbf{m})$ can be expanded as such:

$$p(\mathbf{m}) = p(m_1 m_2 \ldots m_l) \qquad (3)$$

$$= p \begin{pmatrix} \mathfrak{s}_1 & \mathfrak{s}_2 & \cdots & \mathfrak{s}_l \\ \mathfrak{t}_1 & \mathfrak{t}_2 & \cdots & \mathfrak{t}_l \end{pmatrix} \qquad (4)$$

$$\approx p(\mathfrak{s}_1 \mathfrak{s}_2 \ldots \mathfrak{s}_l) + p(\mathfrak{t}_1 \mathfrak{t}_2 \ldots \mathfrak{t}_l) \qquad (5)$$

$$= p_{LM}(\mathfrak{s}_{1:l}) + \sum_{t \in \prod_i t_i} p_{LM}(t_{1:l}) \qquad (6)$$

After expanding the mapping sequences (4), we take an approximation by estimating this as the sum of likelihoods of two sequences formed due to co-segmentations (5). The set of possible TC sequences is given by the Cartesian product of $\mathfrak{t}_i$. These likelihoods can then be estimated using language model (LM) probabilities as shown in (6).

$$\mathbf{t} = \arg\max_{\mathbf{t}' \in \mathbf{m}_t} p(\mathbf{t}') \qquad (7)$$

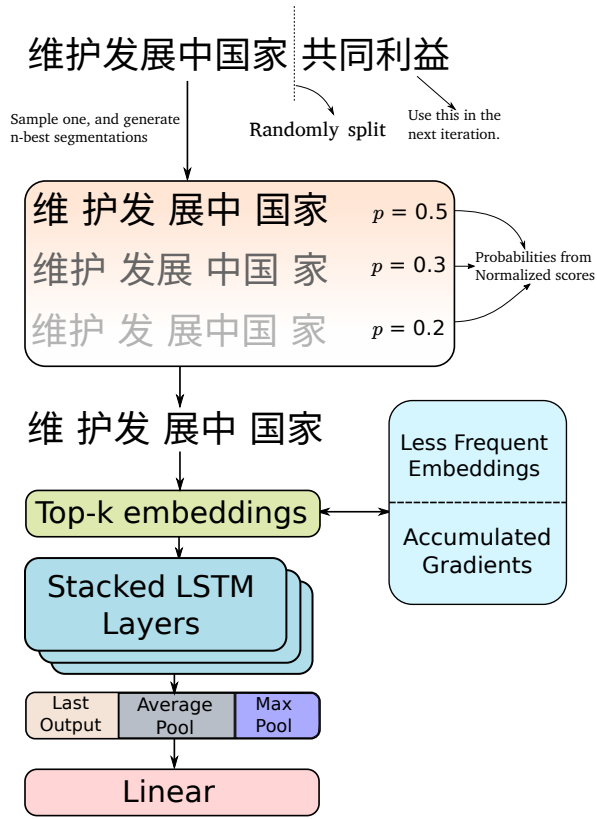

Figure 1: Language model architecture with subword and subsequence sampling. (Alt text: §F.1).

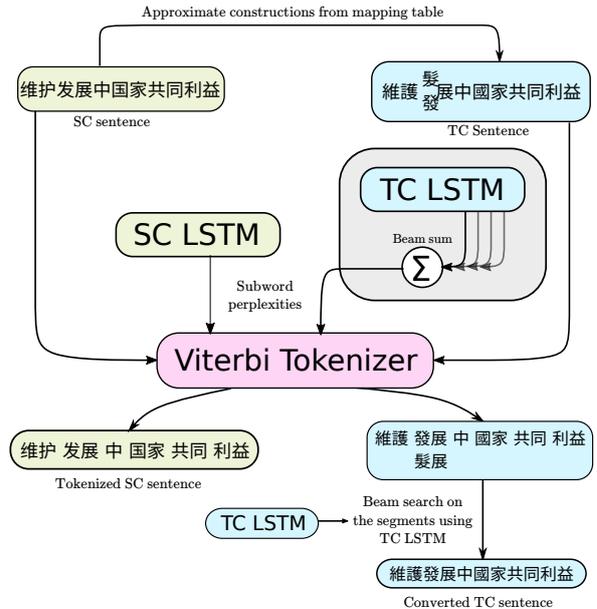

Figure 2: From the given SC sentence, we create possible TC sequences using mappings. We input these to Viterbi, which recursively calls LSTM. Using Eq. (6) as the scoring function, Viterbi outputs the mapping sequence. We perform beam search to find the best TC sequence from the mapping sequence. (Alt text: §F.2).

Once the mapping sequence **m** has been found, all possible TC sequences are found from the set **m_t**, which is the Cartesian product for all $t_i$ in **m**. From (7), we calculate approximate final sequence using beam search.

## 2.2 Model Architecture

Viterbi, a dynamic programming (DP) algorithm, considers phrases (or subsequences) and performs segmentation in a 'bottom-up' fashion (Nagata, 1994; Sproat et al., 1996). RNN-based language models are theoretically considered to be '∞'-gram (Khandelwal et al., 2018), which consitutes a challenge. Consider this sentence, 维护发展中国家共同利益. A potential challenge could be to adequately estimate the probability of 共同利益. As this sequence occurs infrequently in the beginning of sentences in the corpus, an RNN would under-estimate the probability of this subsequence. Moreover, an RNN would likely lose some useful context and perform worse without it (Kim et al., 2019). So for Viterbi to perform well with an RNN, we train the language model on subsequences. We approach this by training our model in such a way that it samples subsequences ran-domly in each epoch. As shown in Fig 1, we randomly split the sentence and use subsequences in separate epochs.

Using Kudo (2018) regularization method, we sample from the $n$-best segmentations in each epoch. This is done so that the model can understand different segmentations of a subsequence using a similar motivation as above. Recent works have shown that varying subword segmentations lead to a better downstream model performance (Provilkov et al., 2019; Kudo, 2018; Hiraoka et al., 2019); therefore, we use it as a data augmentation strategy. Once we get the $n$-best segmentations with scores, we normalize them, and then use the normalized scores as sampling probabilities (see Fig 1). As opposed to other subword tokenizers where the vocabulary size is fixed, we do not limit the vocabulary in our model. Hence, there are numerous possibilities of segment combinations which raises a need of caching most frequent tokens. Inspired by the work related to cache-based LMs (Kawakami et al., 2017) and ghost batches (Hoffer et al., 2017), we only consider the top-$k$ tokens in the main network memory and keep track of gradients of less recently used token embeddings (commonly known as LRU, Least Recently Used policy). This could be thought of as virtual embeddings as delayed gradient accumulation allows to accommodate larger number of tokens. This virtual size embedding architecture is related to the continuous cache implementation and stochastic tokenization architectures (Grave et al., 2016; Hiraoka et al., 2019).

| | HK Literature | HK News | TW Literature | TW News |
|---|---|---|---|---|
| *Sources* | Liu (1962) | Singpao (2017-2018) | Jiubadao (2011) | AS subset Emerson (2005) |
| | Lau Yee (1972) | Mingpao (2017-2018) | Ko (2010) | Liberty Times (2017-2018) |
| | Foon (1988) | CityU subset Emerson (2005) | Yao (1964) | United Daily News (2017-2018) |
| *Average Length* | 194.8 | 214.6 | 188.2 | 223.6 |
| *IAA* | 0.982 | 0.979 | 0.981 | 0.971 |
| *Mapping Examples* | 干 - [幹, 乾, 干, 榦] | 苏 - [蘇, 囌, 甦] | 复 - [復, 複, 覆] | 胡 - [胡, 鬍, 鬍] |
| | 须 - [須, 鬚] | 暗 - [暗, 闇] | 叹 - [嘆, 歎] | 迹 - [跡, 蹟, 迹] |

Table 2: An overview of the dataset used for intrinsic evaluation. We report sources, average character lengths and sentence level inter-annotator agreements (IAA, reported in $\kappa$) and some examples of ambiguous SC-TC mappings.

### 2.3 Segmentation and Disambiguation

This optimal sequencing problem can be formulated as an overlapping subsequence approach, which can be solved using LM based Viterbi (Nagata, 1994; Sproat et al., 1996). Fig. 2 explains this process of joint subword modelling. Here, we take Eq. (6) as the objective function for finding the mapping sequence, however, we use subword perplexities (Cotterell et al., 2018; Mielke et al., 2019; Mielke, 2019) in our implementation. For the TC LSTM, we add the probabilities of the beams of the possible sequences.

As discussed in §2.1 and Eq. (7), beam search is needed to select the best subword sequence for TC. Once the sentences are tokenized, the mapping table is used to convert each SC token to the corresponding TC token. We extract the final TC sentence by resolving ambiguities through beam search using the TC LSTM (Fig. 2).

## 3 Intrinsic Evaluation

### 3.1 Dataset for Intrinsic Evaluation

We construct a gold standard corpus for both Chinese scripts consisting of 4 domains: HK Literature and Newswire, and Taiwanese Literature and Newswire (Table 2) with each domain containing 3000 sentences. SC-TC mapping tables are constructed from existing resources (Denisowski, 2019; Chu et al., 2012). We heuristically convert selected TC sentences to SC using OpenCC. We asked the annotators to manually correct any incorrect conversions.[4]

### 3.2 Language Model Training

We choose the SIGHAN-2005 Bakeoff dataset to train the segmentation-based language model (Emerson, 2005). For SC, we select the PKU and MSR partitions, and for TC, we use the Academia Sinica and CityU partitions. We apply maximal matching (or heuristic dictionary-based word segmenter) to pre-process these datasets by segmenting words into subwords (Wong and Chan, 1996). Here, 'dictionary' refers to the word-list in the mapping table. We then train a 2-layer LSTM language model LSTM with tied weights, and embedding and hidden sizes of 512 (Sundermeyer et al., 2012) on this segmented dataset with subsequence sampling and stochastic tokenization as discussed in §2.2.

### 3.3 Baselines and Ablations[5]

We implement the following baselines for the experimentation:

**Off-the-shelf Converters:** Hanziconv[6] and Mafan[7] are dictionary-based script character converters. Evaluating this could be useful to understand the lower accuracy bound. OpenCC[8] uses a hybrid of characters and words (specifically trie based tokenizer) for script conversion (Pranav A et al., 2019).

**Language Model Disambiguation:** A strong baseline to this problem would be to build a language model to disambiguate between the characters, which is quite similar to STCP (Xu et al., 2017). We use a 2-layer LSTM language model trained on Traditional Chinese corpus.

**Neural Sequence Models:** We heuristically convert Traditional Chinese Wikipedia to Simplified Chinese using OpenCC and use it for training the seq2seq model (Sutskever et al., 2014). We construct a 20-layer neural convolutional sequence model (Gehring et al., 2017) (both in encoder and decoder) using fairseq (Ott et al., 2019).

---

[4] A detailed data statement is given in the appendix.

| Conversion System | HK Lit | | HK News | | TW Lit | | TW News | | Overall | |
|---|---|---|---|---|---|---|---|---|---|---|
| | DED | SA | DED | SA | DED | SA | DED | SA | DED | SA |
| Dictionary based conversion, Hanziconv | 34.1 | 54.7 | 37.7 | 59.1 | 31.3 | 60.0 | 39.3 | 58.9 | 34.2 | 55.6 |
| Dictionary based conversion, Mafan | 14.7 | 71.2 | 17.7 | 72.5 | 14.5 | 73.8 | 13.3 | 72.7 | 14.4 | 72.6 |
| Trie dictionary based conversion, OpenCC | 5.5 | 87.3 | 5.1 | 83.4 | 4.1 | 84.7 | 3.8 | 88.5 | 4.3 | 85.3 |
| Language Model Disambiguation, STCP | 6.3 | 85.6 | 5.4 | 79.9 | 4.7 | 84.1 | 5.2 | 83.9 | 5.3 | 84.0 |
| Convolutional Sequence Models | 6.7 | 85.8 | 5.3 | 79.3 | 4.8 | 84.5 | 5.2 | 83.9 | 5.4 | 84.4 |
| 2kenize with word tokenization | 11.2 | 84.3 | 12.1 | 81.3 | 11.3 | 82.1 | 10.0 | 81.1 | 11.5 | 82.7 |
| 2kenize with maximal matching | 5.2 | 88.7 | _3.3_ | _93.1_ | _4.0_ | _88.6_ | 4.8 | 87.7 | 4.5 | 88.9 |
| 2kenize with Unigram subwords | _3.4_ | _91.9_ | 3.8 | 90.9 | 4.3 | 88.1 | _3.9_ | _87.8_ | _3.7_ | _89.3_ |
| 2kenize with joint LSTM modelling | **2.8** | **94.9** | **3.1** | **93.7** | **3.8** | **91.3** | **2.9** | **91.9** | **3.0** | **92.4** |

Table 3: Results of the intrinsic evaluation experiments which are reported as a mean across 10 different seeds. We use disambiguation error density (DED, the lower, the better) and sentence accuracy (SA, the higher the better) metrics for evaluation. **Bold**: best, Underlined: second-best.

We perform ablation tests by inserting following segmentation models.

**Word tokenization**: We use Jieba, which is a commonly used hidden markov model based word tokenizer for Chinese NLP. [9]

**Dictionary substrings**: We apply maximal string matching, which is a dictionary based greedy tokenizer (Pranav A et al., 2019; Wong and Chan, 1996).

**Unigram** from Sentencepiece: Subword segmentation is performed by sampling unigram language model perplexity values (Kudo, 2018).

**Joint subwords**: As discussed in §2.3, we use joint SC-TC subwords.

### 3.4 Results for Intrinsic Evaluation

We evaluate our models using the metrics of disambiguation error density (DED) and sentence accuracy (SA). DED is the average of total edit distances per 1000 ambiguous Simplified characters, which is $\frac{\sum \text{edit distances}}{\sum \text{ambiguous Simplified characters}} \times 1000$. SA is the number of sentences correctly converted in percentages. Contrary to previous papers, we do not report character based accuracy values, as generally most characters have straightforward mappings — a reason why we opt for a less forgiving metric like SA where every character in a sentence has to be correctly converted.

Results are shown in Table 3, broken down by domain, and overall. Our model attains an average DED of 3.0 and a SA of 92.4% overall, whereas the best existing converter, OpenCC, only achieves a DED of 4.3 and a SA of 85.3%. We find that seq2seq and LM based disambiguation perform almost on par with OpenCC, due to the large number of false positive errors by these models. Jieba achieves an average DED of 11.2 as it does not handle OOV words well. For maximal matching of segmented words and Unigram subwords, it achieves an overall DED of 4.5 and 3.7, respectively — showing that joint segmentation yields better results. We observe that accuracy values are slightly worse on news text, due to the relatively high number of new entities in those datasets. We find that seq2seq and LM based disambiguation gives rise to many false positives. Heuristically converting TC to SC results in certain conversion errors in the training dataset; and additionally, seq2seq approaches tend to reword the target sentence, which shows that they are unsuitable for this task.

### 3.5 Qualitative Error Analysis

We manually inspect incorrect conversions in the intrinsic evaluation and find four interesting recurring linguistic patterns which confused the converters. We instructed the annotators to classify the items in the dataset (overall 12000 sentences in intrinsic evaluation dataset) if the sentences contain any of these patterns. In Table 4, we provide an overview of statistical information of these patterns and the performance by the converters.

**Code mixing**: Vernacular Cantonese characters (*zh-yue*) are a subset of TC characters but do not follow the norms of the standard written Chinese (Snow, 2004). We find that some of the sentences in our dataset are code-mixed with *zh-yue* (e.g. speech transcription) or English (e.g. named entities). Consider the snippet, "... 古惑架 BENZ 190E 撞埋支...", which is code-mixed with

---

| Case | Method | SA | Example | |
|---|---|---|---|---|
| Code mixing with Cantonese (34 cases, 0.3%) | | | 肯尼迪咁多嘢做，掂唔掂呀? | *SC* |
| | | | *With so much to do in Kennedy, can you handle it?* | *HK Lit* |
| | OpenCC | 20.5 | 肯尼迪咁多嘢做，掂唔掂呀? | ✗ |
| | STCP | 8.8 | 肯尼迪咁多嘢做，掂唔掂呀? | ✗ |
| | 2kenize | 91.1 | 甘涹迪咁多嘢做，掂唔掂呀? | ✓ |
| Code mixing with English (1532 cases, 12.8%) | | | 自从我揸住大古惑架 BENZ 190E 揸埋支电灯柱嗰度之后， | *SC* |
| | | | *After I drove Slick's Benz 190E into the telephone pole* | *HK Lit* |
| | OpenCC | 95.6 | 自從我揸住大古惑架 BENZ 190E 揸埋支電燈柱嗰度之後， | ✓ |
| | STCP | 86.5 | 自從我揸住大古惑架 BENZ 190E 揸埋支電燈柱嗰度之後， | ✗ |
| | 2kenize | 98.7 | 自從我揸住大古惑架 BENZ 190E 揸埋支電燈柱嗰度之後， | ✓ |
| Disguised Named Entities (378 cases, 3.15%) | | | 维护发展国家共同利益 | *SC* |
| | | | *Safeguard the common interests of developing countries* | *TW News* |
| | OpenCC | 85.7 | 維護髮展國家共同利益 | ✗ |
| | STCP | 82.1 | 維護髮展國家共同利益 | ✗ |
| | 2kenize | 93.2 | 維護發展國家共同利益 | ✓ |
| Repeated Named Entities (428 cases, 3.57%) | | | 乔治亚来到了乔治亚洲旅游 | *SC* |
| | | | *Georgia came to Georgia for travelling.* | *HK News* |
| | OpenCC | 84.4 | 佐治亞來到了佐治亞洲旅遊 | ✗✗ |
| | STCP | 17.9 | 佐治亞來到了喬治亞洲旅遊 | ✗✓ |
| | 2kenize | 87.8 | 喬治亞來到了喬治亞洲旅遊 | ✓✓ |

Table 4: Casewise breakdown of common errors. The first sentence is SC, second is the English translation and rest are TC outputs from the converters.

both zh-yue and English. The characters "BENZ 190E", 架 and 埋支 are not a part of standard written Chinese. We find that OOV words are *2kenized* into single-character tokens which results in: "古惑 |架 |B|E|N|Z| 1|9|0|E| 撞 |埋 |支" Thus, *2kenize* distributes the entropy over multiple tokens rather than a single token (generally UNK is used in such cases). This allows the language model to have more space for multiple guesses, which shows a massive advantage over word models or just UNK-ing it, a reason why subword tokenizers outperform closed-vocabulary models (Merity, 2019).

**Disguised Named Entities**: Take the recurring sentence: "维护发展国家共同利益". Observe that the sentence contains a frequent word 中国 (China). However, the actual meaning and English translation do not include "China" at all. This is an interesting linguistic trait of Chinese, where words often appear in the sentence, but are not being interpreted. This could easily trip up a tokenizer, as the probability of 中国 being a token independently is high. Having 中国 as a separate token in the sentence could lead into an incorrect conversion (Table 1). We find in *2kenizer*'s trellis[10] that "维护 |发展 | 中" has a higher probability than other possible segmentations. Substructure lookups and beam search in our setup considerably reduces the probability of getting wrong tokenization. The sen-tence is *2kenize*d into "维护 |发展 | 中 | 国家 | 共同 | 利益", which results in the correct conversion – 維護發展中國家共同利益.

**Repetitions**: We find that in 3.57% of sentences, named entities are repeated. Interestingly, STCP, which uses a language model for disambiguation, often only converts one out of the repeated tokens correctly, which we can see in the table. As also shown, STCP prefers 佐治亞 over 喬治亞 in the first occurrence, but then prefers 喬治亞[11] in the second occurrence as it gets more context. *2kenize* converts both of the entities correctly, very likely due to substructure lookups.

**Failure Cases**: Dictionary-based converters (OpenCC, HanziConv and Mafan) only use the first conversion candidate[12] if multiple candidates are available. STCP often converts named entities wrongly, especially the ones which have long-range dependencies and repetitions. Although we find that *2kenize* converts some of the unseen named entities perfectly, some of the errors caused were due to infrequent characters. Few cases are mainly related to variant characters[13] which are often used interchangeably.

---

[10]This is a probability lookup table in Viterbi to keep track of the segment information in a subsequence.

[11]Annotators and sources preferred 喬治亞 over 佐治亞.

[12]We are not sure how did they define the "order", but we have observed that they select more frequent characters as their most highly ranked ones.

[13]For example, 了解 and 瞭解 could be both used for "understand", and 裡面 and 裏面 could be used for "inside".

| Formal Text Classification Dataset Overview | |
| --- | --- |
| *Source* | Singtao |
| *Pretraining Corpus Size* | 17500 |
| *Training Size* | 3000 |
| *Validation Size* | 450 |
| *Testing Size* | 450 |
| *Categories* | Financial, Educational, Local International, Sports |
| *Language* | zh-hant-hk |
| Informal Text Classification Dataset Overview | |
| *Source* | LIHKG |
| *Pretraining Corpus Size* | 21000 |
| *Training Size* | 4000 |
| *Validation Size* | 450 |
| *Testing Size* | 450 |
| *Categories* | Sports, Opinions, IT Financial, Leisure, Memes |
| *Languages* | zh-hant-hk, zh-yue, en-HK |

Table 5: Characteristics of classification dataset (Traditional Chinese) for extrinsic evaluation experiments.

## 4 Extrinsic Evaluation

An accurate script converter should produce a less erroneous dataset, which should in turn improves the accuracy of the downstream tasks. In this section, we demonstrate the effect of script conversion on topic classification tasks to examine this assumption. We also study the impact of tokenization and pooling on the accuracy of topic classification. We apply the converter to the language modelling corpus (Wikitext), then train a classifier for informal and formal topic classification on that translated data. This allows us to measure the performance of the converter compared to other ones for a specific downstream task.

### 4.1 Dataset for Extrinsic Evaluation

This section describes the dataset that we used for extrinsic evaluation experiments. It involves a pretraining dataset which consists Chinese Wikipedia and topic classification datasets.

#### 4.1.1 Pretraining Dataset

We use Chinese Wikipedia articles for pretraining the language model. Script conversion is an issue in Chinese Wikipedia, and currently, they use a server-side mechanism to automatically convert the scripts (dictionary-based) based on the location of the user. However, Wikipedia provides an option to view the article without conversion, which we use in the corpus.[14] We use *zh-CN*, *zh-HK* and *zh-yue* wikis to retrieve articles originally written SC, TC and vernacular Cantonese + TC respectively with the help of wikiextractor[15]. We pretrain the formal text classification models on articles from *zh-HK* and converted *zh-CN*; and classification models for informal text on articles from *zh-HK*, *zh-yue*, and converted *zh-CN*.

#### 4.1.2 Classification Datasets:

We choose two classification tasks: formal news and informal topic classification (Table 5). For formal news, we scrape recent articles (2017-2019) from Singtao[16], for informal topics, we scrape posts (2017-2018) from LIHKG[17].

### 4.2 Performance of various classifiers

For classification baselines, we use character-based SVM (Support Vector Machines, Joachims (1998)), CNN (Convolutional Nets, Zhang et al. (2015)) and Chinese BERT (Devlin et al., 2019). We also employ a state-of-the-art text classifier, MultiFiT (Eisenschlos et al., 2019), a lightweight RNN-based language model based classifier, which has shown to achieve a performance competitive with BERT (Devlin et al., 2019) and ULMFiT (Howard and Ruder, 2018). The base architecture of MultiFiT is a 4-layer QRNN (Bradbury et al., 2016) with classifier head. We choose rectified Adam (Liu et al., 2019) with Lookahead (Zhang et al., 2019) as the optimizer. We employ the cosine cyclic learning scheduler (Smith, 2015), where the limits of learning rate cycles are found by increasing the learning rate logarithmically and computing the evaluation loss for each learning rate (Smith, 2018). To compute the batch size, we apply gradient noise scale to each batch size candidate and pick the one which gives the highest gradient noise scale (McCandlish et al., 2018). We apply label smoothing (Szegedy et al., 2015) and use mixed precision training on RTX 2080. We implement our experiments using Pytorch (Paszke et al., 2019) and FastAI (Howard and Gugger, 2020).

MultiFiT uses *concat pooling* after the last layer of QRNN, which means that the last time step is concatenated with an average and maximum

---

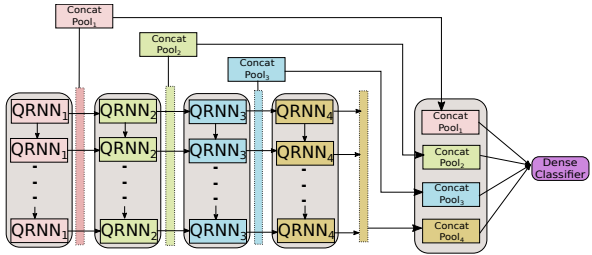

Figure 3: Proposed architecture for topic classification where we tweak MultiFiT to concatenate *concat-pool*s from all layers. (Alt text: §F.3).

| | Formal | Informal |
|---|---|---|
| Char-SVM | 73.2 | 63.7 |
| Char-CNN | 78.5 | 64.9 |
| Chinese BERT (base) | 84.5 | 66.3 |
| MultiFiT with no pooling | 87.5 | 68.5 |
| MultiFiT with concat pooling | <u>88.6</u> | <u>69.9</u> |
| MultiFiT with layer pooling | **89.0** | **70.3** |

Table 6: Performance of various architectures on topic classification in terms of accuracy. The results are reported as a mean result across 10 different seeds and data splits. **Bold**: best, <u>underlined</u>: second best.

pooled over previous time steps. Studies show that in LM based classifiers, different layers capture different types of knowledge–the last layer would be domain-specific and initial layers would be more generalized (Yosinski et al., 2014; Peters et al., 2019). We speculate that *concat pooling* only on the last layer limits the information available to the classifier head and we hypothesise that the classifer would perform better if domain-specific as well as generalized knowledge were available to the head. For this reason, we augment the original MultiFIT architecture with *layer pooling*, which is concat pooling from all the layers, and pass that to the dense layer in the classifier, as shown in Fig 3.

We fine-tune the BERT language model and pretrain the MultiFiT language model on Chinese Wikipedia subsets (§4.1.1). All classifiers are then trained on the given training set (character based models) and evaluated on the test set in terms of accuracy as number of items in each class are roughly equal. This experiment (and subsequent experiments in this section) is repeated across ten different seeds (Reimers and Gurevych, 2018) and data splits (Gorman and Bedrick, 2019) and the results are shown in Table 6. *Layer pooling* shows an absolute improvement of 0.4% improvement over *concat pooling* on formal and informal topic clas-sification, thus confirming our hypothesis.

| Pretraining data of MultiFiT | Formal | Informal |
|---|---|---|
| No Conversions | 89.0 | 70.3 |
| Including conversions with OpenCC | 91.7 | <u>75.6</u> |
| Including conversions with STCP | <u>92.3</u> | 73.4 |
| Including conversions with 2kenize | **93.2** | **77.9** |

Table 7: Ablation test of MultiFiT on different script converters. The results are reported as a mean accuracy result across 10 different seeds and data splits. **Bold**: best, <u>underlined</u>: second best.

| Corpus Tokenization | Formal | Informal |
|---|---|---|
| Char | 93.2 | 77.9 |
| Jieba | 92.4 | 78.3 |
| BPE | 92.7 | 81.0 |
| BPE-Drop | <u>93.7</u> | <u>82.7</u> |
| Unigram | **94.8** | 82.2 |
| 1kenize | **94.8** | **83.2** |

Table 8: Ablation test of MultiFiT on tokenizers. The results are reported as a mean accuracy result across 10 different seeds and data splits. **Bold**: best, <u>underlined</u>: second best.

### 4.3 Effect of Conversion on Classification

For each converter (OpenCC, STCP, *2kenize*), we translate *zh-CN* wiki dataset and augment it with the TC wiki dataset. Then, we pretrain on this dataset, finetune on the domain data and train MultiFiT with layer pooling on these three datasets. We demonstate test set accuracies in Table 7. The dataset translated by *2kenize* outperforms other converters, giving an absolute improvement of 0.9 % on formal and 4.5% over second-best converters on informal topic classification. These results emphasise that better script conversion improves the quality of the pretraining dataset, which boosts the performance of the downstream tasks like topic classification.

### 4.4 Effect of Tokenization on Classification

Studies show that tokenization affects classification accuracy; open-vocabulary methods generally perform best (Eisenschlos et al., 2019; Hiraoka et al., 2019). For this experiment, we perform further ablations on our previous best classifier setup (MultiFiT with layer pooling on *2kenize*) to understand the effect of various subword tokenizers. Pretraining generally takes a long time (1-2 GPU days), hence we pretrain the classifier once for each tokenized corpus and do not perform sub-word sampling for this experiment. For closed vocabulary methods, we use character and word segmentations (here with Jieba). Likewise, for openvocabulary methods, we employ BPE, BPE-Drop and Unigram subword tokenizers.

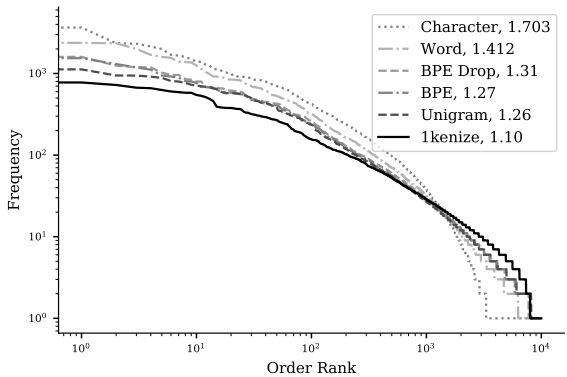

Figure 4: Log-log plots for different tokenizers. This is plotted frequency vs rank for the first 10000 tokens. Negative slopes calcuated from least squares are in the legend (lower means less skewed). (Alt text: §F.4).

Subword tokenizers mostly rely on frequency and do not take likelihood (something similar to $n$-gram language model) of tokenized sentence into consideration. Hence, we choose LM-based Viterbi segmentation (henceforth referred as *1kenize*), and here the LM would be the TC LSTM described in §2.2. We report results in Table 8. We find that for formal classification, *1kenize* and Unigram perform best. *1kenize* outperforms other subword tokenizers for the noisier informal dataset, giving an absolute improvement of 0.5% over the second best method, which is BPE-Drop.

We plot a log frequency of tokens vs log order rank, which is shown in Figure 4. This distribution is based on the LIHKG dataset, which is noisier than other domains. We observe that character and word distributions are steeper than language model based subword tokenizers. This indicates that subword tokenizers produce a less skewed token distribution. Subword tokenizers like BPE and Unigram are deterministic and rely on frequency for segmentation. Since *1kenize* is contextual, being LM-based, we find that it produces the least skewed distribution (lowest Zipf's law coefficient ([Zipf, 1949](#))), which also reduces variance, a reason why this simple segmentation method outperforms others for informal text classification.

## 5 Takeaways and Open Questions

The **contributions** of our work are:

- *2kenize*, a subword segmentation model, which jointly segments source sentence and its corresponding approximate target conversions.
- An **unsupervised script converter** based on *2kenize* which shows a significant improvement over existing script converters and supervised models.
- *1kenize*, a variant of *2kenize* which performs tokenization on only Traditional Chinese sentences which improves accuracy on topic classification tasks.
- **Character conversion evaluation datasets**: spanning Hong Kong and Taiwanese literature and news genres.
- **Traditional Chinese topic Classification datasets**: formal (scraped from Singtao) and informal (scraped from LIHKG) styles spanning genres like news, social media discussions, and memes.

The **key findings** of our work are:

- Our script converter shows a strong performance when dealing with code mixing and named entities. Supervised models are prone to anaphora and unseen entities related errors.
- A simple LM-based Viterbi segmentation model outperforms other subword tokenizers on topic classification tasks and reduces skewness of token distribution on a noisy dataset.

We leave some open questions:

- How can we exploit subword variations to reduce skewness in the NLU tasks?
- Would subword-segmentation-transfer be helpful for other NMT-NLU task pairs like we did for *2kenize* (script conversion) to *1kenize* (classification)?

We anticipate that this study would be useful to TC NLP practitioners, as we address several research gaps, namely script conversion and a lack of benchmark datasets.

## 6 Acknowledgements

The first author would like to thank Dayta AI Limited, S.F. Hui, I-Tsun Cheng, Ishaan Batra, Conrad Ho, Roy Fork, Abhishek Gupta, Ajay Singh, Eugene Ho, Patrick Tu, Alex Chu, and Leland So for making valuable additions to this work. The second author would like to acknowledge funding from the Swedish Research Council for the project under grant agreement 2019-04129, which partly funded this work.

## A  Summary in Traditional Chinese: 簡體中文到繁體中文的文本轉換器

研究中文 NLP 時，將文本進行繁簡轉換是常見的數據預處理步驟。在簡繁轉換過程中，經常出現多個繁字轉換成同一簡體字，反之亦然。藉此透過測試現行的繁簡轉換算法，發現只有 55-85% 準確度。進一步的調查發現，現代的神經網絡，譬如神經語言模型的字符歧義消除 (neural language model character disambiguation) 和神經序列模型 (neural sequence models)，均只達到 84-85% 的句子準確性，都是由第一類錯誤 (Type I error) 所致。我們推斷上述問題，是由於模型未能有效釐清子詞 (subword) 的邊界所導致。

在此，我們提出了 2kenize，一個子詞分割模型 (subword segmentation model)，同時利用先行式繁體中文以及簡體中文進行建構。我們將聯合簡體中文及繁體中文共同訓練 Viterbi 分詞器。即使利用較具挑戰性的數據集測試，本模型亦達到 91-95% 消歧準確度。透過定性誤差分析 (qualitative error analysis)，展示了本模型更擅長處理 code-mixing 以及命名個體 (named entities) 除此以外，我們亦在主題分類領域中進行了外部評估，本模型更在主題分類的字符及詞語模型 (character and word-based models) 的領域中表現出眾，更在子詞正則化 (subword regularization) 中，獲得比 BPE 更好的名次。然後針對繁體中文句子對 2kenize 進行調整，誕生了 1kenize。1kenize 分別在正式數據集與其他子詞分詞器 (subword tokenizers) 名列前茅，在非正式數據集上更表現超群。由此，我們推斷子詞分詞器會嚴重地受 token 的分佈及偏度而影響

是次研究的貢獻：

1. 2kenize：簡體中文到繁體中文的文本轉換器

2. 字符轉換評估數據集：跨越香港和台灣文獻及新聞等多個類型的數據集

3. 主題分類數據集：繁體中文的正式和非正式文本數據涵蓋新聞，社交媒體討論，改圖，改歌，memes 等二次創作文本。

## B  Data Statement for Intrinsic Evaluation

### B.1  Corpus

In this subsection, we discuss the annotation procedure and the characteristics of the corpus used for the intrinsic evaluation. We have used Bender and Friedman (2018) data statement design for the description.

### B.1.1  Curation Rationale

The script conversion task is understudied in NLP and we could not find good quality parallel corpora to evaluate our approaches. The idea is to curate a diverse collection of TC works and convert them to SC, due to its one-to-one correspondence. However, we find out that some of the conversions were wrong because

1. sometimes dictionaries resulted in incorrect conversion,
2. stylistic differences between HK and TW characters and phrasing,
3. code-mixing of Cantonese and Traditional Chinese,
4. code-mixing with non-Chinese characters,
5. some characters in TC-SC conversion have one-to-many mappings as well.

Hence, we need quality control with human annotators to validate our conversions.

### B.1.2  Annotation Process

**Demographic**: We opted for 4 trained annotators, 2 for annotating HK-style TC and 2 for annotating TW-style TC and thus going for double annotation for the corpus. They ranged in age from $18-20$ years, included 2 men and 2 women, gave their ethnicity as Hong Kongers (2) and Taiwanese (2), and their native spoken languages were Cantonese (2) and Taiwanese Mandarin (2).

**Workload**: Annotators approximately validated 100 sentences per hour, comprising of total workload of 60 hours. They were given a month to annotate and were paid 5000 Hong Kong Dollars on completion.

**Procedure**: The annotators were shown TC and converted SC sentences (we used OpenCC to convert) and were asked to validate and correct any conversion mistakes. In case of disagreement, we used majority voting between automatically converted and annotators' corrections.

We provide raw agreement and Krippendorf's $\alpha$ in Table 1 for pooled data and various sub-groups of the dataset. We also report inter-annotator agreements on character and phrasal levels in Table 2. These agreement values are difficult to interpret, but generally $\alpha \geq 0.8$ is considered to be substantial.

|        |      | RA    | $\alpha$ |
|--------|------|-------|------|
| **HK** |      | 0.98  | 0.98 |
|        | Lit  | 0.982 | 0.98 |
|        | News | 0.979 | 0.97 |
| **TW** |      | 0.98  | 0.98 |
|        | Lit  | 0.981 | 0.98 |
|        | News | 0.971 | 0.97 |

Table 9: Inter-annotator agreements

|                 | RA   | $\alpha$ |
|-----------------|------|------|
| Character Level | 0.98 | 0.97 |
| Word Level      | 0.95 | 0.94 |
| Sentence Level  | 0.93 | 0.92 |

Table 10: Inter-annotator agreements as per different levels

### B.1.3 Speech Situation

The publication dates and sources are listed in the Table 2. HK and TW literature consists of popular books for which many movie and drama adaptations are made.[18] Specifically, for HK literature, the text contains code-mixed characters with Vernacular Cantonese, which is quite unusual in formal publishing practices, and these books are often cited as an example for popularizing Cantonese in the 60s (Snow, 2004). We also found code-mixing with English and numerous transliterated named entities which we have used for qualitative error analysis in the Table 4.

### B.1.4 Text Characteristics

Although Hong Kong and Taiwan both use Traditional Chinese, they are stylistically different as the dominant spoken language in HK is Cantonese and in TW is Taiwanese Mandarin. Thus, it is quite essential to test the performance of our algorithms on these two styles. We collected two genres for each style: informal literature and formal news. We found more variation within informal HK-TW literature as compared to the formal news. We intentionally chose long sentences (average length of 200 characters), especially which contain more ambiguous characters to make the dataset more challenging for testing.

## C  Data Statement for Extrinsic Evaluation

This subsection describes the characteristics of the topic classification in Traditional Chinese. For the short overview, please see Table 5.

### C.1  Curation Rationale

We choose two different styles for curating this dataset: formal and informal. The formal text consists of news dataset scraped from Singtao, one of the popular newswire in Hong Kong. The classes in this dataset consist of Financial, Educational, Local, International, and Sports subsections. There are 17500 unlabelled and 3900 labelled items in this section. Authors would like to credit I-Tsun Cheng for giving us helpful suggestions in curating this dataset.

The informal text consists of social media posts dataset scraped from LIHKG, a Twitter equivalent in Hong Kong. The classes in this dataset consist of Sports, Opinions, Memes, IT, Financial and Leisure. There are 21000 unlabelled and 4900 labelled items in this section. Authors would like to credit Leland So for giving us helpful suggestions in curating this dataset.

### C.2  Language Variety

The texts in the formal subsection are typically written in Hong Kong style Traditional Chinese (zh-hant-hk). The posts scraped from LIHKG are predominantly in Traditional Chinese (zh-hant-hk), and they are often code-mixed with Vernacular Cantonese (zh-yue) and English (en-HK).

### C.3  Speaker Demographic

Speakers were not directly approached for inclusion in this dataset and thus could not be asked for demographic information. Our best guess for demographic of LIHKG forum users are typically university students (19-23 years), and the majority of them speak Cantonese as a native language.

### C.4  Text Characteristics

The news articles are scraped from 2017-2019 and LIHKG posts are scraped from 2017-2018. Some of the posts in LIHKG are in the transliterated Cantonese form and some of them are not written in Standard Written Chinese. The news posts are generally quite long and often contains more than 5 sentences (average length of nearly 300 characters). On the other hand, the LIHKG posts are shorter and forums titles are generally one sentence each (average length of nearly 50 characters). Please note that due to the current situations in Hong Kong, we do not include political posts and news from mid-2019.

---

[18] We highly recommend these movies and novels as well.

## D Description of Intrinsic Evaluation Experiments

### D.1 Heuristic Grid Search of Learning Rate and Batch Size Hyperparameters

We employ the cosine cyclic learning scheduler (Smith, 2015), where the limits of learning rate cycles are found by increasing the learning rate logarithmically and computing the evaluation loss for each learning rate (Smith, 2018). To compute the batch size, we apply gradient noise scale to each batch size candidate and pick the one which gives the highest gradient noise scale (McCandlish et al., 2018).

### D.2 Training of SC and TC Language Model

The datasets are described in §3.2. The model architecture is 2-layer LSTM language model with tied weights. Embedding size is 512 and hidden size is 512. We perform a concat pooling in the last layer where we concatenate the last output of the word, mean pool and max pool of all representations. We adopt comparable subword perplexity as suggested by Cotterell et al. (2018); Mielke et al. (2019); Mielke (2019), where we use a common denominator, referring to the number of segments per word in order to compare. On average, we achieve a perplexity of 168.6 on the Chinese Treebank test set (Nianwen et al., 2016). Also refer to Chinese LM Benchmark: `https://chinesenlp.xyz/#/docs/language_modeling`. The training took 2 days on RTX 2080 with FP16 training, with a batch size of 256 and number of epochs of 250.

### D.3 Training of Convolutional seq2seq

Training dataset is a heuristically converted Traditional Chinese Wikipedia with OpenCC. We use 20 layers in encoder and decoder with the embedding size of 512 implemented in Fairseq (Ott et al., 2019). Dropout is 0.1 and we use adaptive softmax to speed up the training. The training took 1 day on RTX 2080 with FP16 training, with a batch size of 128 and number of epochs of 250.

## E Description of Extrinsic Evaluation Experiments

### E.1 Character CNN training

The datasets are described in §4.1.2. The model architecture is 7-layer CNN with tied weights and residual blocks. Embedding size is 512 and hidden size is 512. We perform a concat pooling in the last layer where we concatenate the last output of the word, mean pool and max pool of all representations. The training took 16 hours on RTX 2080 with FP16 training, with a batch size of 256 and number of epochs of 350.

### E.2 Chinese BERT training

The datasets are described in §4.1.2. We use Chinese BERT base (12-layer, 768-hidden, 12-heads, 110M parameters) using Transformers library (Wolf et al., 2019). We use sequence length of 384 and batch size of 12. Finetuning language model took 2 hours (learning rate of 3e-5) and finetuning classifier took 1 hour each on both datasets, including grid search on learning rates: 3e-4, 1e-4, 5e-5, 3e-5, where 3e-5 gives the best results (on RTX 2080 with FP16 training).

### E.3 MultiFiT training

We found MultiFiT is highly reproducible as compared to other models as it gives the least variance across the seeds and data splits. Hyperparameters are chosen by heuristic grid search on learning rate and batch size. The datasets are described in §4.1.2. Pretraining language model takes 1 GPU day for each experiment of MultiFiT. Finetuning language model takes 3 hours where we used a patience of 2 epochs. Finetuning classifiers takes 3 hours where we used a patience of 2 epochs. All experiments of MultiFiT are implemented using FastAI (Howard and Gugger, 2020).

## F Alternative texts for figures and Chinese explanations

### F.1 Alternative text for Figure 1

The recurring Chinese sentence is split and we take one subsequence of it. The other subsequence is used in next iteration. We perform Unigram viterbi segmentation on this and get the probabilities. The probabilities are normalized and we sample a segmentation using this probability. This segmentation goes into the model which goes through cached embeddings, followed by stacked LSTM layers, followed by concat pooling (which consists of last output, mean pooling and max pooling) which then goes through a linear layer. We cache the top-k embeddings in the main memory and for the least frequent embeddings we track the gradients and do not keep them in the main network (we used gradient accumulation).

### F.2 Alternative text for Figure 2

From the given SC sentence, we create possible TC sequences using mappings. We input these to Viterbi, which recursively calls LSTM. Using Eq. (6) as the scoring function, Viterbi outputs the mapping sequence. We perform beam search to find the best TC sequence from the mapping sequence where we used the same TC LSTM again.

### F.3 Alternative text for Figure 3

The architecture contains 4 stacked QRNN layers. Each layer has QRNN cells. After every layer we perform a concat pool (taking the last output, max pool and mean pool). We aggregate these pools in the final layer which goes into a linear layer. We highly recommend this for making the training more stable.

### F.4 Alternative text for Figure 4

We have plotted log-log token distribution. On x-axis we have order rank and on y-axis we have frequencies. Character based tokenization gives a slope of 1.703, BPE-Drop gives 1.31, BPE gives 1.27, word tokenization (Jieba) gives 1.41, unigram sampling gives 1.28 and 1kenize gives the least skewed distribution with a slope of 1.1. Note that these are negative slope and lower the slope is, more efficiently vocabulary is tokenized.

### F.5 Recurring Chinese sentence

Here, we explain the recurring sentence in this paper. In Table 1 we had SC sentence 维护发展中国家共同利益, which means *Safeguarding the common interests of developing countries*. This is pronounced as *wéi hù fā zhǎn zhōng guó jiā gòng tóng lì* in Mandarin. Its correct TC translation is 維護發展中國家共同利益, which is pronounced as *wai4 wu6 faat3 zin2 zung1 gwok3 gaa1 gung6 tung4 lei6 jik1* in Cantonese (note that the numerals are the tones).